\documentclass{llncs}
\usepackage[dvips]{graphicx}
\usepackage[scriptsize,bf]{caption}
\usepackage{doublespace}
\newcommand{\normalstretch}{0.999}
\begin{document}

\title{Market-Based Reinforcement Learning in Partially Observable
  Worlds}
\author{Ivo Kwee, Marcus Hutter \and J\"{u}rgen Schmidhuber }
\institute{IDSIA, Manno CH-6928, Switzerland. \\
  \email{\{ivo,marcus,juergen\}@idsia.ch}
  \thanks{This work was supported by SNF grants 21-55409.98 and
    2000-61847.00 }
}

\maketitle              

\abstract{Unlike traditional reinforcement learning (RL), market-based
  RL is in principle applicable to worlds described by partially
  observable Markov Decision Processes (POMDPs), where an agent needs
  to learn short-term memories of relevant previous events in order to
  execute optimal actions.  Most previous work, however, has focused
  on reactive settings (MDPs) instead of POMDPs.  Here we reimplement
  a recent approach to market-based RL and for the first time evaluate
  it in a toy POMDP setting.  }


\section{Introduction}

One major reason for the importance of methods that learn sequential,
event-memorizing algorithms is this: sensory information is usually
insufficient to infer the environment's state ({\em perceptual
  aliasing}, Whitehead 1992).  This complicates goal-directed
behavior.  For instance, suppose your instructions for the way to the
station are: ``Follow this road to the traffic light, turn left,
follow that road to the next traffic light, turn right, there you
are.''  Suppose you reach one of the traffic lights. To do the right
thing you need to know whether it is the first or the second.  This
requires at least one bit of memory --- your current environmental
input by itself is not sufficient.

Any learning algorithm supposed to {\em learn} algorithms that store
relevant bits of information from training examples that only tell
whether the current attempt has been successful or not --- this is the
typical reinforcement learning (RL) situation --- will face a major
temporal credit assignment problem: which of the many past inputs are
relevant (and should be represented in some sort of internal memory),
which are not?
Most traditional work on RL requires Markovian interfaces to the
environment: the current input must provide all information about
probabilities of possible inputs to be observed after the next action,
e.g., \cite{Sutton:88,WatkinsDayan:92}.  These approaches, however,
essentially learn how to react in response to a given input, but
cannot learn to identify and memorize important past events in
complex, partially observable settings, such as in the introductory
example above.

Several recent, non-standard RL approaches in principle are able to
deal with partially observable environments and can learn to memorize
certain types of relevant events
\cite{Kaelbling:95,Littman:95,Jaakkola:95,Cliff:94,%
McCallum:93,Ring:94,Schmidhuber:91nips,Wiering:96levin,Wiering:97ab,%
Schmidhuber:97bias}.  None of them, however, represents a satisfactory
solution to the general problem of learning in worlds described by
{\em partially observable Markov Decision Processes} (POMDPs).
%
The approaches are either ad hoc, or work in restricted domains only,
suffer from problems concerning state space exploration versus
exploitation, have non-optimal learning rate or have limited
generalization capabilities. Most of these problems are overcome
in~\cite{hutter:01aixi} but the model is yet computationally
intractable.
%
Here we consider a novel approach to POMDP-RL called {\em market-based
  RL} which in principle does not suffer from the limitations of
traditional RL.  We first give an overview of this approach and its
history, then evaluate it in a POMDP setting, and discuss its
potential and limitations.

\section{Market-based RL: History \& State of the Art}

\paragraph{Classifier systems.}
Probably the first market-based approach to RL is embodied by
Holland's Classifier Systems and the Bucket Brigade algorithm
\cite{Holland:85}.  Messages in form of bitstrings of size $n$ can be
placed on a global message list either by the environment or by
entities called classifiers. Each classifier consists of a condition
part and an action part defining a message it might send to the
message list. Both parts are strings out of $\{0,1,\_ \}^{n}$ where
the `\_' serves as a `don't care' if it appears in the condition part.
A non-negative real number is associated with each classifier
indicating its `strength'.  During one cycle all messages on the
message list are compared with the condition parts of all classifiers
of the system. Each matching classifier computes a `bid' based on its
strength.  The highest bidding classifiers may place their message on
the message list of the next cycle, but they have to pay with their
bid which is distributed among the classifiers active during the last
time step which set up the triggering conditions (this explains the
name bucket brigade).  Certain messages result in an action within the
environment (like moving a robot one step). Because some of these
actions may be regarded as 'useful' by an external critic who can give
payoff by increasing the strengths of the currently active
classifiers, learning may take place.  The central idea is that
classifiers which are not active when the environment gives payoff but
which had an important role for setting the stage for directly
rewarded classifiers can earn credit by participating in `bucket
brigade chains'.  The success of some active classifier recursively
depends on the success of classifiers that are active at the following
time ticks. Bankrupt classifiers are removed and replaced by freshly
generated ones endowed with an initial amount of money.

\paragraph{PSALMs.}
Holland's approach suffers from certain drawbacks though.  For
instance, there is no credit conservation law --- money can be
generated out of nothing.  Pages 23-51 of \cite{Schmidhuber:87} are
devoted to a more general approach called {\em Prototypical
  Self-referential Associating Learning Mechanisms} (PSALMs).
Competing/cooperating agents bid for executing actions. Winners may
receive external reward for achieving goals.  Agents are supposed to
learn the credit assignment process itself (meta-learning). For this
purpose they can execute actions for collectively constructing and
connecting and modifying agents and for transferring credit (reward)
to agents.  PSALMs are the first systems that enforce the important
constraint of total credit conservation (except for consumption and
external reward) - this constraint is not enforced in Holland's
classifier economy, which may cause money inflation and other
problems. PSALMs also inspired the money-conserving {\em Neural
  bucket brigade}, where the money is "weight substance" of a
reinforcement learning neural net \cite{Schmidhuber:89cs}.

\paragraph{Hayek machines} \cite{Baum:99}
are designed to avoid further loopholes in Holland's credit assignment
process that may allow some agent to profit from actions that are not
beneficial for the system as a whole.  Property rights of agents are
strictly enforced.  The current owner of the right to act in the world
computes the minimal price for this right, and sells it to the highest
bidder. Agents can create children and invest part of their money into
them, and profit from their children's success. Wealth and bid of an
agent are separate quantities. All of this makes Hayek more stable
than the original bucket brigade.  There have been impressive
applications --- Hayek learned to solve complex blocks world problems
\cite{Baum:99}.  Hayek is reminiscent of PSALM3 \cite{Schmidhuber:87},
but a crucial difference between PSALM3 and Hayek may be that PSALM3
does not strictly enforce individual property rights.  For instance,
agents may steal money from other agents and temporally use it in a
way that does not contribute to the system's overall progress.

\section{The Hayek4 System}

We briefly describe our reimplementation of Hayek4~\cite{Baum:2000}
--- the most recent Hayek machine variant. Unlike earlier Hayek
versions, Hayek4 uses a new representation language.

\paragraph{Economic model.}
Hayek4 consists of a collection of agents that each have a
rule, a possible action, a wealth and a numerical bid. The agents are
required to solve an instance of a complex task. But because each
agent can only perform a single action, they need to cooperate.
In a single instance, computation proceeds in a series of
\emph{auctions}.  The highest bidding agent wins and ``owns'' the
world, pays an amount equal to its bid to the previous owner and may
then perform its action on the world. The owner of the world collects
any reward from it, plus the payment of the next owner. Only if the
actual revenue is larger than its bid, an agent can increase its
\emph{wealth}.

\paragraph{Birth, death and taxes.}
Agents are allowed to create \emph{children} if their wealth is larger
than some predetermined number (here 1000). Parents endorse their
offspring with a minimum amount of money which is subtracted from
their wealth, but in return parents receive a share $c$ of their
children's future profit. Children's rules are randomly created,
copied or mutated from their parents with probabilities $p_r$, $p_c$
and $p_m$, respectively. See Table~\ref{tbl:parameters} for their
numerical values.

By the definition of wealth, birth processes automatically focus on
rather successful agents. To improve efficiency, after each instance
all agents pay a small amount of \emph{tax} proportional to the amount
of computation time they have used. To remove unuseful agents, earlier
Hayek versions required agents to pay a small fixed amount of tax;
agents without money were removed from the system. Hayek4 removes them
once they have been inactive for the last, say, 100 instances.

\paragraph{Post production system.}
The agents in Hayek4 are essentially a \emph{Post} production system.
Baum and Durdanovic argue that this provides a richer representation
than the S-expressions in earlier Hayek versions because Post systems
are Turing complete (like the representation used by PSALMs
\cite{Schmidhuber:87}).  Each agent forms a Post rule of the form $L
\rightarrow R$ where $L$ is the antecedent (or rule) and $R$ is the
consequent (the action).  The state is encoded in a string $S$; the
agent is only allowed to perform $R$ when $L$ matches string $S$.
This procedure is iterated until no rules match, and computation
halts.

\section{Implementation}

We found Hayek4 rather difficult to reproduce. Several parameters
require tuning --- making Hayek stable still seems to remain an art.
Parameter settings of our Hayek system are summarized in
Table~\ref{tbl:parameters}; for a detailed explanation
see~\cite{Baum:2000}. For small task sizes our program tends to find
reliable solutions, but larger sizes cause stability problems, or
prevent Hayek4 from finding a solution within several days.
\begin{table}[tb] \center
\scriptsize \begin{tabular}{lll}
    symbol      & value & description \\ \hline
    $W$         & 10.0  & Baum and Durdanovic's $W$ parameter \\
    $R$         & 100.0 & reward value \\
    $\epsilon$  & 0.01  & children always bid this amount higher than highest bid \\
    $u$         & 0.25  & increase level if (1-$u$) instances are solved;
                          decrease if less than $u$. \\
    $t$         & 1e-3  & execution tax \\
    $c$         & 0.25  & copyright share \\
    $m$         & 0.5   & mutation ratio \\
    $p_r$       & 0.3   & probability of random rule for new child \\
    $p_c$       & 0.3   & probability of copying rule for new child \\
    $p_m$       & 0.3   & probability of mutation of rule for new child \\
\end{tabular}
\caption{Parameter settings for our Hayek4 reimplementation. For
  a detailed explanation see~\cite{Baum:2000}.}
\label{tbl:parameters}
\end{table}

We tried to reimplement Hayek4 as closely as possible, except
for certain deviations justified by improved performance or
implementation issues:
\begin{itemize}
\item {\em Single rule:} instead of agents having multiple rules, we
  allowed only one rule per agent.
\item {\em Fixed reward:} instead of paying off reward proportional to
  the size of the task, we used a fixed reward of 100.
\item {\em TD backup:} we additionally applied time differential
  (TD)~\cite{Sutton:88} backup of the agent's bid, as traditionally
  done in reinforcement learning systems, to accelerate convergence.
\end{itemize}

\paragraph{Blockworld.}

Baum and Durdanovic reported a set of Hayek agents that collectively
form a universal solver for arbitrary blockworld problems
\cite{Baum:2000} involving arbitrarily high stacks of blocks. The set
consists of five rules listed in the table below.

\newcommand{\ra}{$\rightarrow$}

\begin{figure}[h]
\begin{minipage}[b]{0.48\textwidth}
\centering \includegraphics[width=18mm]{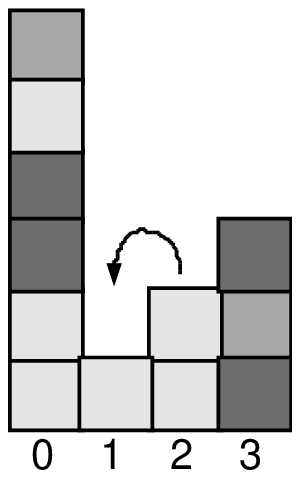}
\caption{Example of Blockworld instance with stack height=6. The
  objective is to replicate in stack 1 the color sequence represented by the blocks
  in stack 0, by moving blocks between stacks 1-3. Here the optimal next move would be to move the top
      block of stack 2 to stack 1. }
\label{fig:blockworld}
\end{minipage} \hfill
\begin{minipage}[b]{0.48\textwidth}
  \centering {\scriptsize \tt
\begin{tabular}{lllll}
             &~~~& \multicolumn{3}{c}{agent}    \\ \cline{3-5}
world        & & rule         & action & bid   \\ \hline
babc:cbb::a  & & *:*:*:*      & 1\ra3  & 7.78  \\
babc:cb::ab  & & *:*:*:*      & 1\ra3  & 7.78  \\
babc:c::abb  & & *:*:*:*      & 1\ra3  & 7.78  \\
babc:::abbc  & & *.*:1:*:*2*  & 3\ra2  & 8.07  \\
babc::c:abb  & & *.*:1:*:*2*  & 3\ra2  & 8.07  \\
babc::cb:ab  & & *.*:1:*2:*   & 2\ra1  & 8.07  \\
babc:b:c:ab  & & *.*:1:*:*2*  & 3\ra2  & 8.07  \\
babc:b:cb:a  & & *.*:1:*:*2*  & 3\ra2  & 8.07  \\
babc:b:cba:  & & *.*:1:*2:*   & 2\ra1  & 8.07  \\
babc:ba:cb:  & & *.*:1:*2:*   & 2\ra1  & 8.07  \\
babc:bab:c:  & & *.:1:*2:*    & 2\ra1  & 35.80 \\
babc:babc::  & & &
\end{tabular}}
\caption{A typical Blockworld solution for stack height=4. Each
  colon-separated field of the world string represents a stack of
  blocks of colors {\tt \{a,b,c\}}. The agent's rule matches the world
  string; ``*'' represents a ``don't care'' symbol; numbers in the
  rule encode replacements of matched substrings of stack 0. Actions
  move blocks between stacks 1, 2 and 3. }
\label{fig:universal}
\end{minipage}
\end{figure}

We tested Hayek4 by first hardwiring a system consisting of those five
so-called universal agents only. Indeed, it reliably generated stacks
of increasing size, reaching stack level 20 within 800 instances, and
level 30 within 1000 instances. Figure~\ref{fig:universal} shows a
typical run resulting in a stack of height=4. Then we switched on the
creation of children but kept mutation turned off.  This caused Hayek
to progress at a much slower pace, probably because new children
disrupt the universal solver as they always win the auctions using the
$\epsilon$-bid scheme.

\paragraph{Problems.}
Unfortunately, our system was not able to find the universal solver by
itself. It typically reached stack size 5, but then failed to progress
any further. We believe that with increasing task size the system is
encountering stability problems. Much remains to be done here.

\section{Adding Memory to Hayek}
Unlike traditional RL, market-based RL is in principle applicable to
POMDPs.  Previous work, however, has usually focused on reactive
settings (MDPs) instead of POMDPs (see \cite{Wilson:94} for a notable
exception though).  Here we add a memory register to the Hayek machine
and apply it to POMDPs. The memory register acts as a additional
environmental variable that can be manipulated by the agent using
\emph{write} actions.

\begin{figure}[t] \centering
  \includegraphics[height=22mm]{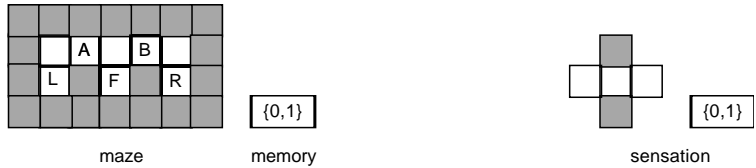}
\caption{
  Woods101 is a partially observable environment; grey positions
  represent walls. Left: The agent starts at one of the positions $L$
  or $R$ and has to reach the food at $F$.  The two aliased positions
  are $A$ and $B$. A 1-bit memory register is available for
  writing/reading.  Right: the agent senses the four neigbouring
  positions and the value of the memory register; here, for example,
  the agent might be either at $A$ or at $B$.}
\label{fig:woods101}
\end{figure}

\paragraph{Test problem.}
To study Hayek4's performance on POMDPs, we focus on one of the
simplest possible POMDP problems. Figure~\ref{fig:woods101} shows our
toy example which is an adapted version of Woods101~\cite{Cliff:94}, a
standard test case for non-Markov search problems. Agents start in
either of the two positions, $L$ or $R$, and have to reach the food at
$F$. Agents have only limited perception of their four immediate
neighbouring positions \emph{north}, \emph{east}, \emph{south} and
\emph{west} plus the value of the memory register. Without memory,
there is no Markov solution to this problem because there are two
aliased positions $A$ and $B$ with identical environmental input where
the optimal agent needs to execute different actions: in $A$ the
optimal action is \emph{east}; in $B$ the optimal action is
\emph{west}. Note that the starting positions $L$ and $R$ are also
aliased. However, in both $L$ and $R$ the optimal action is
coincidentally the same (i.e. going \emph{north}).

\paragraph{Results.}
We ran Hayek4 on our test problem. After about 1000 instances it
learned a policy that solved the POMDP problem using a 1-bit memory.
In early instances the number of agents grew up to 1000 or more but
finally only 8 agents remained. While the bids of all agents converged
to the actual reward value of 100, their wealth values varied
considerably. Sample histories of solved instances starting at $L$ are
shown in Table~\ref{tbl:policy-L}; histories of solved instances
starting at $R$ are shown in Table~\ref{tbl:policy-R}.

\begin{table}[t] \quad
\begin{minipage}[b]{0.5\textwidth} \scriptsize
\begin{tabular}{lllclcrr}
\multicolumn{3}{c}{state} &~~~&  \multicolumn{4}{c}{agent} \\
\cline{1-3} \cline{5-8}
x & y~~ & m && rule      & action & bid     & wealth \\ \hline
1 & 1 & 0 && 1:0:0:0:0 & m1  & 100.00 & 119.51 \\
1 & 1 & 1 && 1:*:*:0:1 & n   & 100.00 & 184.04 \\
1 & 2 & 1 && *:*:1:0:1 & m0  & 100.00 & 301.70 \\
1 & 2 & 0 && 0:1:1:0:0 & e   & 100.00 & 116.76 \\
2 & 2 & 0 && 0:1:0:1:0 & e   & 100.00 & 951.08 \\
3 & 2 & 0 && 0:1:1:1:* & s   & 100.00 & 103.52 \\
3 & 1 & 0 &&  &  &  & \\
\end{tabular}
\end{minipage}
\quad
\begin{minipage}[b]{0.5\textwidth} \scriptsize
\begin{tabular}{lllclcrr}
\multicolumn{3}{c}{state} &~~~&  \multicolumn{4}{c}{agent} \\
\cline{1-3} \cline{5-8}
x & y~~ & m && rule      & action & bid     & wealth \\ \hline
1 & 1 & 1 &&  1:*:*:0:1 & n   &  100.00  & 185.01 \\
1 & 2 & 1 &&  *:*:1:0:1 & m0  &  100.00  & 302.15 \\
1 & 2 & 0 &&  0:1:1:0:0 & e   &  100.00  & 117.22 \\
2 & 2 & 0 &&  0:1:0:1:0 & e   &  100.00  & 951.40 \\
3 & 2 & 0 &&  0:1:1:1:* & s   &  100.00  & 103.51 \\
3 & 1 & 0 &&  &  &  & \\
  &   &   &&  &  &  & \\
\end{tabular}
\end{minipage}
\caption{Two policies from starting position $L$=(1,1); auctions proceed
  from the first line downwards. Left: the memory is
  set to $m$=1 before going \emph{north} but again reset to $m$=0 before
  the aliased position (2,2) is reached. Right: the memory is
  initially set to $m$=0 by a memory agent at (1,2). The first four fields
  of the rule represent the bit values of \emph{north},
 \emph{east}, \emph{south} and \emph{west}, respectively; the fifth field
 matches the value of the memory register. Star ('*') variables represent
 ``don't care'' symbols.}
\label{tbl:policy-L}
\end{table}

\begin{table}[t] \quad
\begin{minipage}[b]{0.5\textwidth} \scriptsize
\begin{tabular}{lllclcrr}
\multicolumn{3}{c}{state} &~~&  \multicolumn{4}{c}{agent} \\
\cline{1-3} \cline{5-8}
x & y~~ & m && rule      & action & bid     & wealth \\ \hline
5 & 1 & 1 && 1:*:*:0:1 & n      & 100.00  & 183.80 \\
5 & 2 & 1 && 0:0:1:1:1 & w      & 100.00  & 143.04 \\
4 & 2 & 1 && 0:1:0:*:1 & w      & 100.00  & 378.54 \\
3 & 2 & 1 && 0:1:1:1:* & s      & 100.00  & 103.52 \\
3 & 1 & 1 &&  &  &  & \\
  &   &   &&  &  &  & \\
\end{tabular}
\end{minipage}
\quad
\begin{minipage}[b]{0.5\textwidth} \scriptsize
\begin{tabular}{lllclcrr}
\multicolumn{3}{c}{state} &~~&  \multicolumn{4}{c}{agent} \\
\cline{1-3} \cline{5-8}
x & y~~ & m && rule      & action & bid     & wealth \\ \hline
5 & 1 & 0 && 1:0:0:0:0 & m1 & 100.00 &  120.21 \\
5 & 2 & 1 && 1:*:*:0:1 & n  & 100.00 &  185.33 \\
5 & 2 & 1 && 0:0:1:1:1 & w  & 100.00 &  143.75 \\
4 & 2 & 1 && 0:1:0:*:1 & w  & 100.00 &  379.05 \\
3 & 2 & 1 && 0:1:1:1:* & s  & 100.00 &  103.50 \\
3 & 1 & 1 && &  &  & \\
\end{tabular}
\end{minipage}
\caption{Two policies from starting position $R$=(5,1); auctions proceed
  from the first line downwards. Left: the memory is initially set to
  $m$=1 and the agent proceeds directly \emph{north} to $F$=(5,2).
  Right: the memory was initially set to $m$=0 and a memory agent sets
  $m$=1 before proceeding \emph{north}.}
\label{tbl:policy-R}
\end{table}

The results can be summarized as follows. When an instance starts from
$L$, Hayek assures that the memory is set to 0 before it reaches
aliased position $A$. If the memory was initially set to $m$=1, a
write agent sets the memory to 0; if the memory was already $m$=0,
nothing is done. When the agent starts from the right, the opposite
occurs: before Hayek reaches aliased position $B$, it checks if the
memory bit is properly set to $m$=1. Now, two different agents ---
with identical perception of the environment but different perception
of the memory --- act at aliased positions $A$ and $B$: an \emph{east}
agent at $A$ if the memory bit is 0, and a \emph{west} agent if the
memory is set to $m$=1.

Hayek also evolves one agent executing action \emph{south} at sequence
ends at (3,2). It disregards the value of the memory by using the
`*'-symbol (``don't care'').  ``Don't care'' symbols are useful to
prevent unnecessary o\-ver-spe\-cia\-li\-zation, i.e., we do not need
\emph{two} separate agents for each memory value.  In fact, such
over-specialization occurred with the starting agent going
\emph{north}: Hayek did not use a ``don't care'' symbol for the memory
register and therefore needed a extra memory agent ($m1$) if the bit
was initially unset.

\paragraph{Discussion.}
This simple test problem shows that Hayek, in principle, is able to
solve POMDPs. More precisely, using a memory register,
Hayek ``de-aliases'' aliased positions in the POMDP by tagging these
situations \emph{before} they actually occur. We note that the
solution found by Hayek is not unique: it does not matter
whether it chooses to write a ``0'' or a ``1'' for the left path, as
long as the right path uses the complement. Once a policy seems to
perform well, Hayek converges toward this local optimum and sticks to
it.

Note that the found solution is nearly but not exactly optimal; an
extra $m1$-memory agent was necessary for the left policy in
Table~\ref{tbl:policy-L} because Hayek was not able to generalize
optimally in case of the agent going \emph{north}, failing to use a
``don't care'' symbol for the memory value.

\section{Conclusion}

We have started to evaluate market-based RL in POMDP settings,
focusing on the Hayek machine as a vehicle for learning to memorize
relevant events in short-term memory. Using a memory bit register,
Hayek was able to distinguish aliased positions in a toy POMDP and
evolve a stable solution. The approach is promising, yet much remains
to be done to make it scalable.

%

\end{document}